\documentclass[11pt]{article}
\pdfoutput=1
\usepackage{acl2016}
\usepackage{times}
\usepackage{url}
\usepackage{latexsym}
\usepackage{textcomp}
\usepackage{graphicx}
\usepackage{caption}
\usepackage{subcaption}
\usepackage[caption=false,labelformat=simple]{subfig}
\usepackage{adjustbox}
\usepackage{amsmath}
\usepackage{makecell}

\DeclareMathOperator*{\argmax}{\arg\!\max}
\usepackage{verbatim}
\usepackage{amssymb}

\aclfinalcopy 
\def\aclpaperid{407} 

\title{Automatic Open Knowledge Acquisition via Long Short-Term Memory Networks with Feedback Negative Sampling}

\author{Byungsoo Kim \quad\quad\quad\quad Hwanjo Yu \quad\quad\quad\quad Gary Geunbae Lee \\\\
Department of Computer Science and Engineering \\
POSTECH (Pohang University of Science and Technology) \\
Pohang, Republic of Korea \\
{\tt \{bsmail90, hwanjoyu, gblee\}@postech.ac.kr}}

\date{}

\begin{document}
\maketitle
\begin{abstract}
Previous studies in Open Information Extraction (Open IE) are mainly based on extraction patterns. They manually define patterns or automatically learn them from a large corpus. However, these approaches are limited when grasping the context of a sentence, and they fail to capture implicit relations. In this paper, we address this problem with the following methods. First, we exploit long short-term memory (LSTM) networks to extract higher-level features along the shortest dependency paths, connecting headwords of relations and arguments. The path-level features from LSTM networks provide useful clues regarding contextual information and the validity of arguments. Second, we constructed samples to train LSTM networks without the need for manual labeling. In particular, feedback negative sampling picks highly negative samples among non-positive samples through a model trained with positive samples. The experimental results show that our approach produces more precise and abundant extractions than state-of-the-art open IE systems. To the best of our knowledge, this is the first work to apply deep learning to Open IE.
\end{abstract}

\section{Introduction}
Open Information Extraction (Open IE) is a task that involves taking sentences and extracting the arguments and the relations between them. Open IE systems extract this information in the form of a triple or n-tuple. Consider the following input sentence: \textit{`Boeing announced the 747 ASB in 1986'}. An Open IE system will extract \textit{\textless Boeing; announced; the 747 ASB\textgreater} and \textit{\textless Boeing; announced the 747 ASB; in 1986\textgreater}, or \textit{\textless Boeing; announced; the 747 ASB; in 1986\textgreater}. Open IE has been successfully applied in many NLP tasks, such as question answering~\cite{OQA}, knowledge base (KB) population~\cite{KBP}, and ontology extension~\cite{OpenIESyntacticSemantic}. The major difference between traditional IE and Open IE is domain dependency. Traditional IE requires a pre-defined set of relations, whereas Open IE~\cite{Textrunner} does not. Open IE represents relations with the words in a sentence. This new paradigm removes domain dependency, extending the relation set to whole word-sets. Thus, it is possible to run Open IE at the scale of the Web.

Previous Open IE systems adopt two main approaches. The first approach involves manually defining the extraction patterns to find the relationships between arguments. Reverb~\cite{Reverb} showed that simple parts-of-speech (POS) patterns can cover the majority of relationships.~\newcite{DependencyOpenIE} and KRAKEN~\cite{KrakeN} manually define extraction rules in dependency parse trees. The second approach involves automatically learning a set of dependency-based extraction patterns from a large corpus. Methods adopting this second approach include WOE~\cite{WOE}, OLLIE~\cite{OLLIE}, and ReNoun~\cite{ReNoun}.

Although previous Open IE systems have been used in many other studies, these systems only extract relations that are represented explicitly in a sentence. For example, previous systems find the (explicit) relation of \textit{`capital'} between \textit{`Vilnius'} and \textit{`Lithuania'} in the following sentences: \textit{`The two countries were officially at war over \textbf{Vilnius}, the \textbf{capital} of \textbf{Lithuania}'}; \textit{`The geographical midpoint of Europe is just north of \textbf{Lithuania}\textquotesingle s \textbf{capital}, \textbf{Vilnius}'}; and \textit{`\textbf{Vilnius} was the \textbf{capital} of \textbf{Lithuania}, the residence of the Grand Duke'}. The explicit relations accompany text snippets, which are strong clues regarding the relation (\textit{`\textbf{Vilnius}, the \textbf{capital} of \textbf{Lithuania}'}, \textit{`\textbf{Lithuania}\textquotesingle s \textbf{capital}, \textbf{Vilnius}'}, and \textit{`\textbf{Vilnius} was the \textbf{capital} of \textbf{Lithuania}'}). However, previous Open IE systems fail to find the relation when it is implicitly represented in a sentence, such as \textit{`He returned to \textbf{Lithuania} and then lived in the \textbf{capital}, \textbf{Vilnius}, until his death'}. Unlike explicit relations, an implicit relation is not captured merely with textual patterns. Extracting these implicit relations involves a deeper understanding of the context of a sentence.

In this paper, we propose a novel Open IE system that automatically extracts features using long short-term memory (LSTM) networks. The bi-directional recurrent architecture with LSTM units automatically extracts higher-level features along the shortest dependency paths connecting headwords of relations and arguments. Because these paths contain only informative words that are relevant to finding the proper arguments of the relation, the extracted features can grasp contextual information without superfluous information. Because there are no prevalent datasets for training Open IE systems, we propose methods for constructing training samples. In particular, feedback negative sampling selects highly negative samples among non-positive samples, and decreases disagreements between positive and negative samples. The procedure for constructing the training set is fully automatic. It does not require any manual labeling. The experimental results show that our proposed system produces 1.62 to 4.32 times more correct extractions, including implicit relations, with higher precision than state-of-the-art Open IE systems.

The remainder of this paper is organized as follows. Section \ref{sec:types_of_relation} describes the two types of relations that our system aims to extract. Section \ref{sec:task_definition} defines Open IE as two tasks: argument detection and preposition classification. Section \ref{sec:automatically_constructing_the_training_set} describes the procedure for automatically constructing the training set. Sections \ref{sec:argument_detection} and \ref{sec:preposition_classification} provide detailed explanations of the neural network architectures for argument detection and preposition classification, respectively. Section \ref{sec:triple_extraction} describes how triples are extracted from the outputs in argument detection and preposition classification. Section \ref{sec:experiments} describes experimental settings and shows evaluation results. Finally, Section \ref{sec:conclusion} concludes our work.

\section{Types of Relation}
\label{sec:types_of_relation}
The first type of relation is a verb-mediated relation. A relation of this type is a verb phrase. It often forms an n-ary relation. Consider as an example: \textit{`Boeing announced the 747 ASB in 1986'}. The relation \textit{`announced'} has 3 arguments: \textit{`Boeing'}, \textit{`the 747 ASB'}, and \textit{`1986'}. This n-ary relation is represented as an n-tuple: \textit{\textless Boeing; announced; the 747 ASB; in 1986\textgreater}. However, because a binary relation is a core concept of the semantic web and ontological KB, the n-ary relation must be converted to binary relations. This conversion involves handling the problem of incomplete relations. In the above example, by merely spanning all pairs of arguments, the triples are \textit{\textless Boeing; announced; the 747 ASB\textgreater}, \textit{\textless Boeing; announced in; 1986\textgreater}, and \textit{\textless the 747 ASB; announced in; 1986\textgreater}. However, the relation \textit{\textless Boeing; announced in; 1986\textgreater} omits critical information---namely, \textit{`the 747 ASB'}---and fails to find the complete relation between \textit{`Boeing'} and \textit{`1986'}. The appropriate triple, without loss of information, is \textit{\textless Boeing; announced the 747 ASB in; 1986\textgreater}. Because \textit{`the 747 ASB'} is a patient of \textit{`announce'} in the case of \textit{\textless the 747 ASB; announced in; 1986\textgreater}, the appropriate triple is \textit{\textless the 747 ASB; be announced in; 1986\textgreater}. Note that we do not restore the complete passive form (\textit{`was announced in'}). Rather, \textit{`be announced in'} is sufficient for indicating the passive form and for downward application.

Another type of relation is a noun-mediated relation. A relation of this type is a noun phrase. As described in~\newcite{ReNoun}, a noun-mediated relation is an attribute of an argument. Consider as an example: \textit{`He sat on the board of Meadows Bank, an independent bank in Nevada'}. A triple with a noun-mediated relation is \textit{\textless Meadows Bank; an independent bank in; Nevada\textgreater}. In this triple, \textit{`Nevada'} is the target of an attribute, \textit{`an independent bank'}, and \textit{`Meadows Bank'} is the value of the attribute. We add \textit{`be'} to the relation phrase in order to specify its meaning as an attribute, resulting in \textit{\textless Meadows Bank; be an independent bank in; Nevada\textgreater}. Unlike verb-mediated relations, the conversion from a noun-mediated n-ary relation to binary relations merely involves spanning all pairs of arguments.

\section{Task Definition}
\label{sec:task_definition}
We define Open IE as two tasks: argument detection, and preposition classification. Given a sentence, detecting the argument involves regarding a certain word (\textit{rel}) as a headword of a relation and then classifying other words (\textit{arg}) as to whether they are the proper headwords of the arguments for that relation. As an input, the classifier takes the shortest dependency path connecting \textit{rel} to \textit{arg}. We denote this path as \textit{path(rel, arg)}. By considering the shortest dependency path connecting two words, we can concentrate on informative words that are useful for understanding the relation between the two words~\cite{ShortestDependencyPath}.
\begin{figure}[t]
    \centering
    \begin{subfigure}[b]{0.237\textwidth}
        \includegraphics[width=\textwidth]{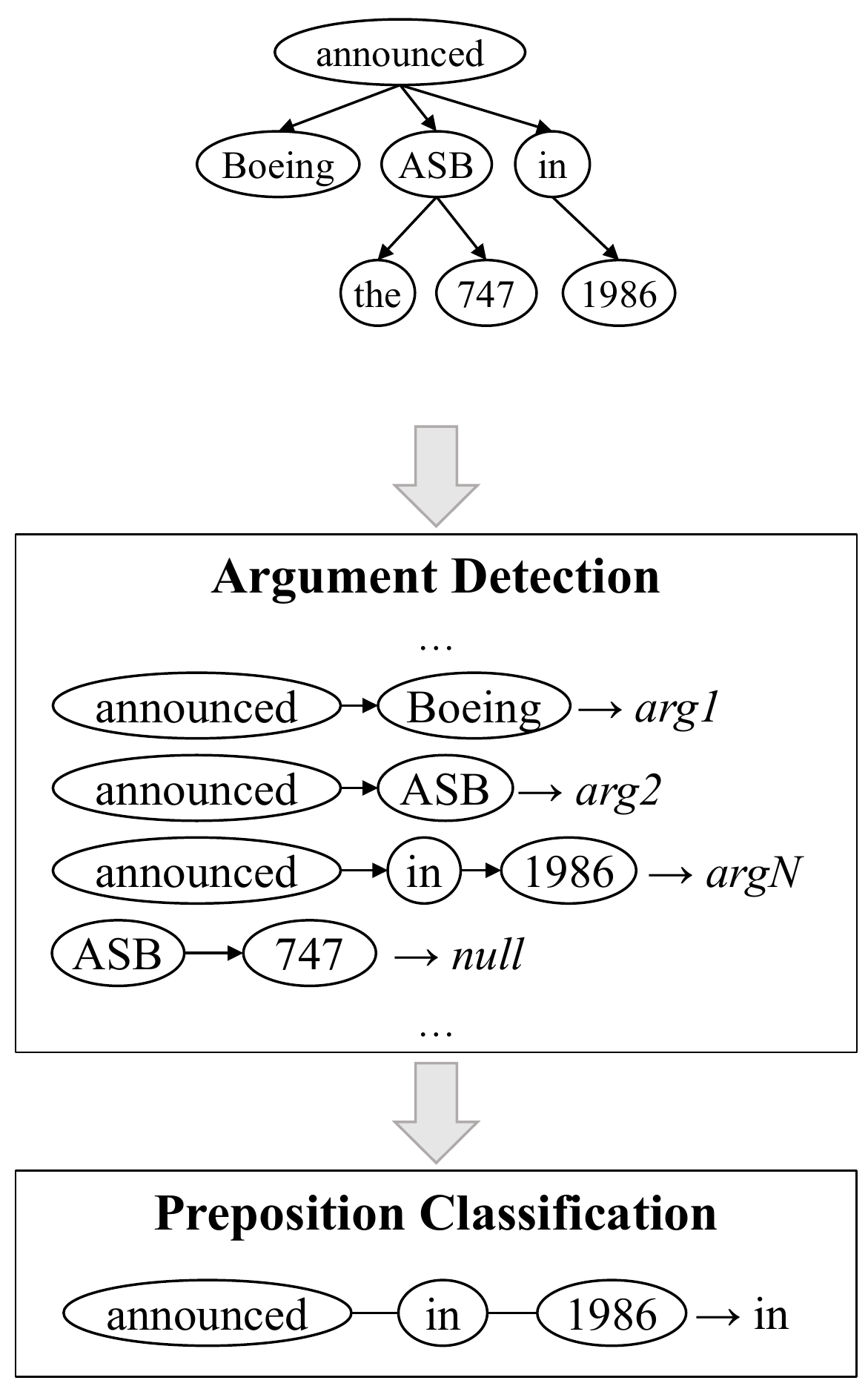}
        \caption{}
        \label{fig:task_definition_verb}
    \end{subfigure}
    \begin{subfigure}[b]{0.238\textwidth}
        \includegraphics[width=\textwidth]{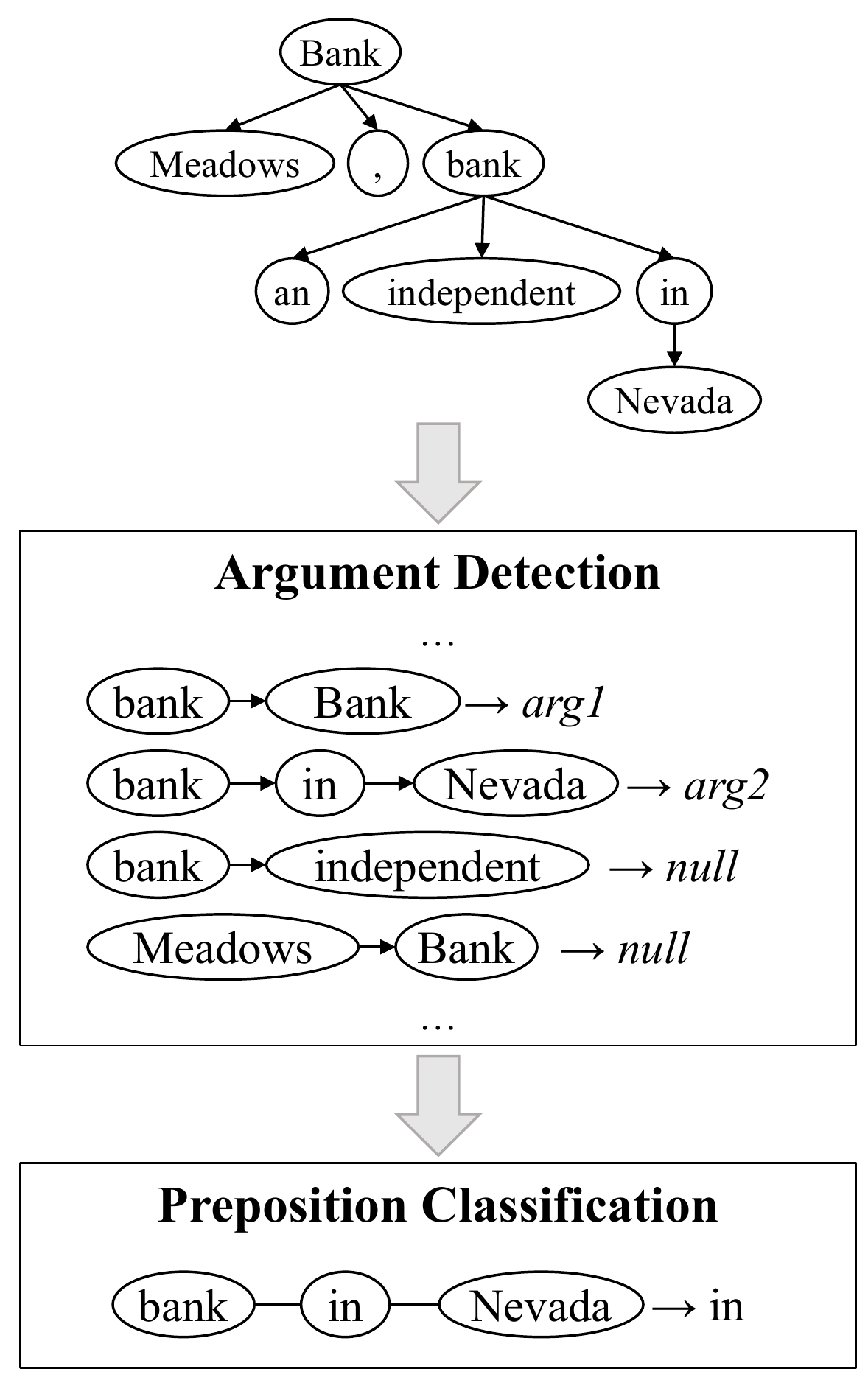}
        \caption{}
        \label{fig:task_definition_noun}
    \end{subfigure}
    \caption{Argument detection and preposition classification, given the following sentences: (a) \textit{`Boeing announced the 747 ASB in 1986'}, and (b) \textit{`Meadows Bank, an independent bank in Nevada'}. For simplicity, we do not specify the dependency relations.}
    \label{fig:task_definition}
\end{figure}
For example, in Figure \ref{fig:task_definition_verb}, \textit{`Boeing'}, \textit{`ASB'}, \textit{`the'}, and \textit{`747'} are irrelevant for determining whether \textit{`1986'} is a proper argument for \textit{`announced'}. We define four classes for argument detection: \textit{arg1}, \textit{arg2}, \textit{argN}, and \textit{null}. For a verb-mediated relation, \textit{arg1} and \textit{arg2} are the agent and patient of a relation, respectively, and \textit{argN} denotes other arguments. For a noun-mediated relation, \textit{arg1} and \textit{arg2} are the value and target of a relation, respectively. We do not classify \textit{argN} in the case of a noun-mediated relation. Finally, \textit{null} denotes a term that is not an argument. In Figure \ref{fig:task_definition_verb}, argument detection classifies \textit{path(announced, Boeing)}, \textit{path(announced, ASB)}, and \textit{path(announced, 1986)} as \textit{arg1}, \textit{arg2}, and \textit{argN}, respectively. Other paths are classified as \textit{null}. If the argument detection classifies \textit{path(rel, arg)} with the verb \textit{rel} as \textit{argN} or the noun \textit{rel} as \textit{arg2}, the preposition classification finds the appropriate preposition between \textit{rel} and \textit{arg}. In Figure \ref{fig:task_definition_verb}, preposition classification selects \textit{`in'} as the appropriate preposition between \textit{`announced'} and \textit{`1986'}.

\section{Automatically Constructing the Training Set}
\label{sec:automatically_constructing_the_training_set}
\subsection{Highly Precise Tuple Extraction}
As~\newcite{SRLOpenIE} leveraged semantic role labeling (SRL) to find n-ary relations, we used SRL\footnote{We used ClearNLP (\url{www.clearnlp.com}) for the natural language processing pipeline.} to extract highly precise tuples with verb-mediated relations\footnote{We used the English Wikipedia corpus to construct the training set.}. We assign \textit{rel} to \textit{predicate}, and \textit{arg1}, \textit{arg2}, and \textit{argN} to the labeled word with the roles \textit{A0}, \textit{A1}, and \textit{AM}, respectively. If the word is a preposition, we apply the assignment to its child, while retaining the lemma of the preposition for preposition classification. Consider the following example: \textit{`In addition to the French Open, Nadal won 10 other singles titles in 2005'}. The SRL output is \textit{`predicate: win, A0: Nadal, A1: titles, AM-DIS: In, AM-TMP: in'}, and our assignment extracts the tuple as \textit{`rel: win, arg1: Nadal, arg2: titles, argN (in): addition, argN (in): 2005'}. To minimize tuple-extraction errors, we only extract tuples from the top 1M sentences with the highest SRL confidence scores.

\begin{figure}[t]
    \centering
    \includegraphics[width=0.45\textwidth]{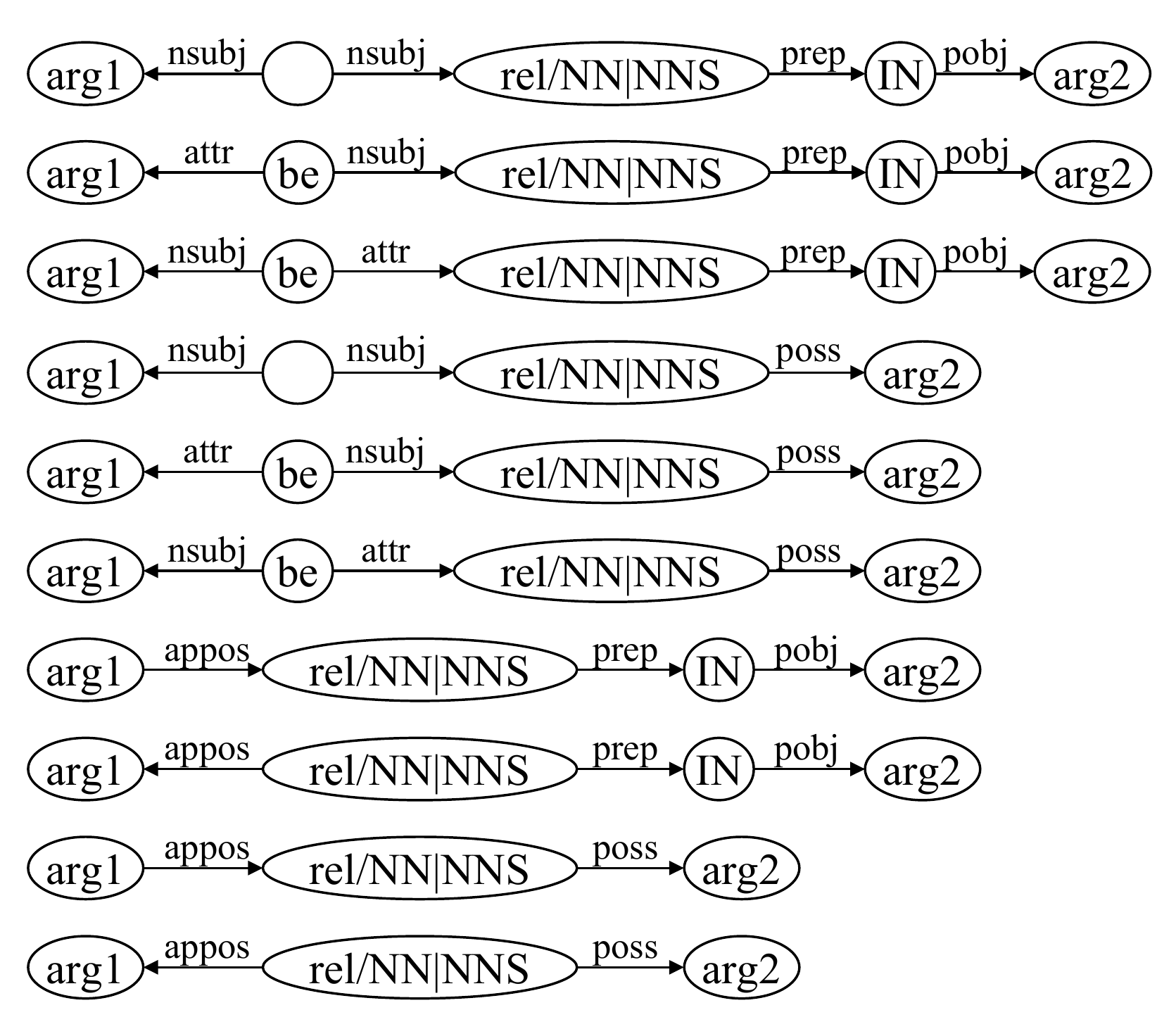}
    \caption{We restrict the POS tag of the relation to \textit{NN} or \textit{NNS}. Empty circles mean that we do not give any constraints. Circles with \textit{IN} and \textit{be} restrict the POS tag and lemma of the words to \textit{IN} and \textit{`be'}, respectively.}
    \label{fig:noun_seed_patterns}
\end{figure}

We define ten dependency-based extraction patterns to extract highly precise tuples with noun-mediated relations (see Figure \ref{fig:noun_seed_patterns}). The patterns are applied to subgraphs of a dependency parse tree. The circles and arrows in the patterns represent words and dependency relations in the subgraph, respectively. For the preposition classification, we retain a lemma of a word at a circle with \textit{IN}. If there is no circle with \textit{IN} in a pattern, we retain \textit{`of'}. If a pattern is matched, we assign \textit{arg1}, \textit{rel}, and \textit{arg2} to the words in the circles with \textit{arg1}, \textit{rel}, and \textit{arg2}, respectively. For example, from the dependency parse tree of the sentence, \textit{`The agency is located in Gaborone, capital of Botswana'}, the seventh pattern in Figure \ref{fig:noun_seed_patterns} is matched and our assignment extracts the tuple as \textit{`rel: capital, arg1: Gaborone, arg2 (of): Botswana'}. Like verb-mediated tuple extraction, we only extract tuples from the top 1M sentences with the highest dependency-parsing confidence scores.

\subsection{Training Set Augmentation}
The goal of training set augmentation is to find sentences representing relations in highly precise tuples that SRL and the patterns failed to capture. Similar to OLLIE~\cite{OLLIE} and ReNoun~\cite{ReNoun}, this augmentation process is based on seed-based distant supervision: if arguments in a seed triple appear in a sentence, their relation is likely to appear in the sentence. The augmentation begins by converting the tuple to triples. For tuples with a verb-mediated relation, we convert each tuple to \textit{\textless arg1; rel; arg2\textgreater}, \textit{\textless arg1; rel; argN\textgreater}, and \textit{\textless arg2; rel; argN\textgreater}. Tuples with a noun-mediated relation are converted to \textit{\textless arg1; rel; arg2\textgreater}. Among the converted triples, we acquire 55K seeds satisfying the following constraints: (1) the arguments are proper nouns or cardinal numbers; (2) arguments with a proper noun are properly linked to entities in DBpedia~\cite{DBpedia}; and (3) the lemma of a relation is not \textit{`be'} or \textit{`do'}. We use DBpedia Spotlight~\cite{DBpediaSpotlight} for entity linking. For each seed triple, we find sentences containing the same linked entities of arguments with a proper noun or the same surface forms of arguments with a cardinal number. Because the distant supervision hypothesis is often erroneous, we include the following constraints: (1) the sentence contains the lemma of a relation; (2) the headwords of relations and arguments are connected via a linear dependency path; and (3) triples with verb-mediated relations have a path length of less than seven. For example, we acquired the seed triple from the tuple, \textit{`rel: capital, arg1: Gaborone, arg2 (of): Botswana'} with its arguments linked to DBpedia entities, \textit{`Gaborone'} and \textit{`Botswana'}. We retrieved the corpus and found \textit{`Now Prime Minister of Bechuanaland, Khama continued to push for \textbf{Botswana}\textquotesingle s independence, from the newly established \textbf{capital} of \textbf{Gaborone}'}. The augmentation produces 110K (sentence, seed triple) pairs that cannot be covered by highly precise tuples. We label \textit{path(rel, arg1)}, \textit{path(rel, arg2)}, and \textit{path(rel, argN)} from these pairs and highly precise tuples as \textit{arg1}, \textit{arg2}, and \textit{argN}, respectively. These labeled paths comprise positive samples for argument detection. We also label \textit{path(rel, argN)} and \textit{path(rel, arg2)} with the noun \textit{rel} as their prepositions to comprise samples for preposition classification.

\subsection{Feedback Negative Sampling}
Samples from the previous stages merely indicate which paths are \textit{arg1}, \textit{arg2}, and \textit{argN}. They do not describe which paths are \textit{null} (negative). One possible option for negative sampling is to regard non-positive paths as negative ones. However, this risks treating uncaptured positive paths as negative ones.
\begin{figure}[t]
    \centering
    \includegraphics[width=0.45\textwidth]{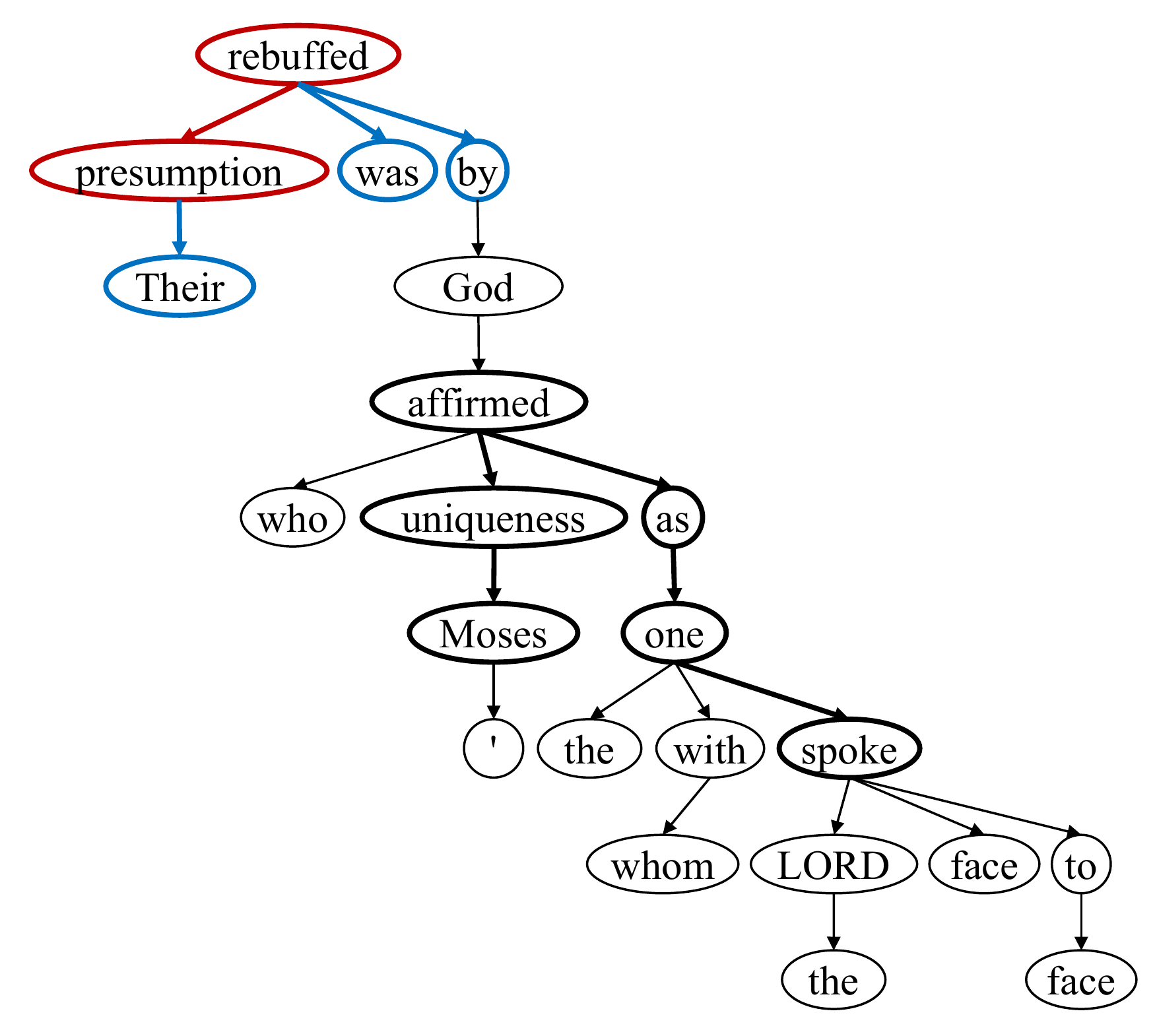}
    \caption{Feedback negative sampling acquires highly negative paths, such as \textit{path(rebuffed, Their)}, \textit{path(was, presumption)}, and \textit{path(by, presumption)}, without mistakenly treating uncaptured positive paths, such as \textit{path(spoke, Moses)}, as negative paths.}
    \label{fig:feedback_negative_sampling}
\end{figure}
For example, we found that \textit{path(spoke, Moses)} is an uncaptured positive path with the label \textit{argN} in the sentence: \textit{`Their presumption was rebuffed by God who affirmed Moses\textquotesingle~uniqueness as the one with whom the LORD spoke face to face'} (see Figure \ref{fig:feedback_negative_sampling}). Our strategy for addressing this problem begins from the observation that there are two features to a highly negative path: (1) it contains a positive path, or a positive path contains it; and (2) the more similar it is to the positive path, the more negative the path is. For example, in Figure \ref{fig:feedback_negative_sampling}, \textit{path(rebuffed, their)}, \textit{path(was, presumption)}, and \textit{path(by, presumption)} are highly negative paths. They contain \textit{path(rebuffed, presumption)}, which is positive, and they have only one more node than the positive path. From this observation, we describe feedback negative sampling in Algorithm 1.
\begin{table}[t]
    \centering
    \begin{adjustbox}{width=0.47\textwidth}
    \begin{tabular}{l}
        \Xhline{3\arrayrulewidth}
        \textbf{Algorithm 1:} Feedback negative sampling \\
        \Xhline{3\arrayrulewidth}
        \textbf{Input:} $NP$ = A set of non-positive paths \\
        $N$ = An empty set of negative samples \\
        $F$ = A model trained on positive samples \\
        $p$ = A prediction score threshold \\
        \textbf{foreach} non-positive path $np \in NP$ \textbf{do} \\
        \quad $F(np)^i$ = Prediction score of $np$ on the $i$-th class \\
        \quad $m$ = $\argmax_{i}\limits F(np)^i$ \\
        \quad \textbf{if} $F(np)^m > p$ \textbf{then} \\
        \quad \quad \textit{N} = \textit{N} $\cup$ \{\textit{np}\} \\
        \textbf{return} \textit{N} \\
        \Xhline{3\arrayrulewidth}
    \end{tabular}
    \end{adjustbox}
\end{table}
The rationale behind this algorithm is as follows: because a model\footnote{We describe the details for this model in the next section.} trained with positive samples assigns high confidence to positive paths, it also assigns high confidence to non-positive paths with these features. For each non-positive path, we obtain clues (feedback) regarding these features. If the path has these features, the model assigns a high prediction score to a certain class. We regard the path as a negative sample when the score exceeds a certain threshold.

\section{Argument Detection}
\label{sec:argument_detection}
\begin{figure}[t]
    \centering
    \includegraphics[width=0.45\textwidth]{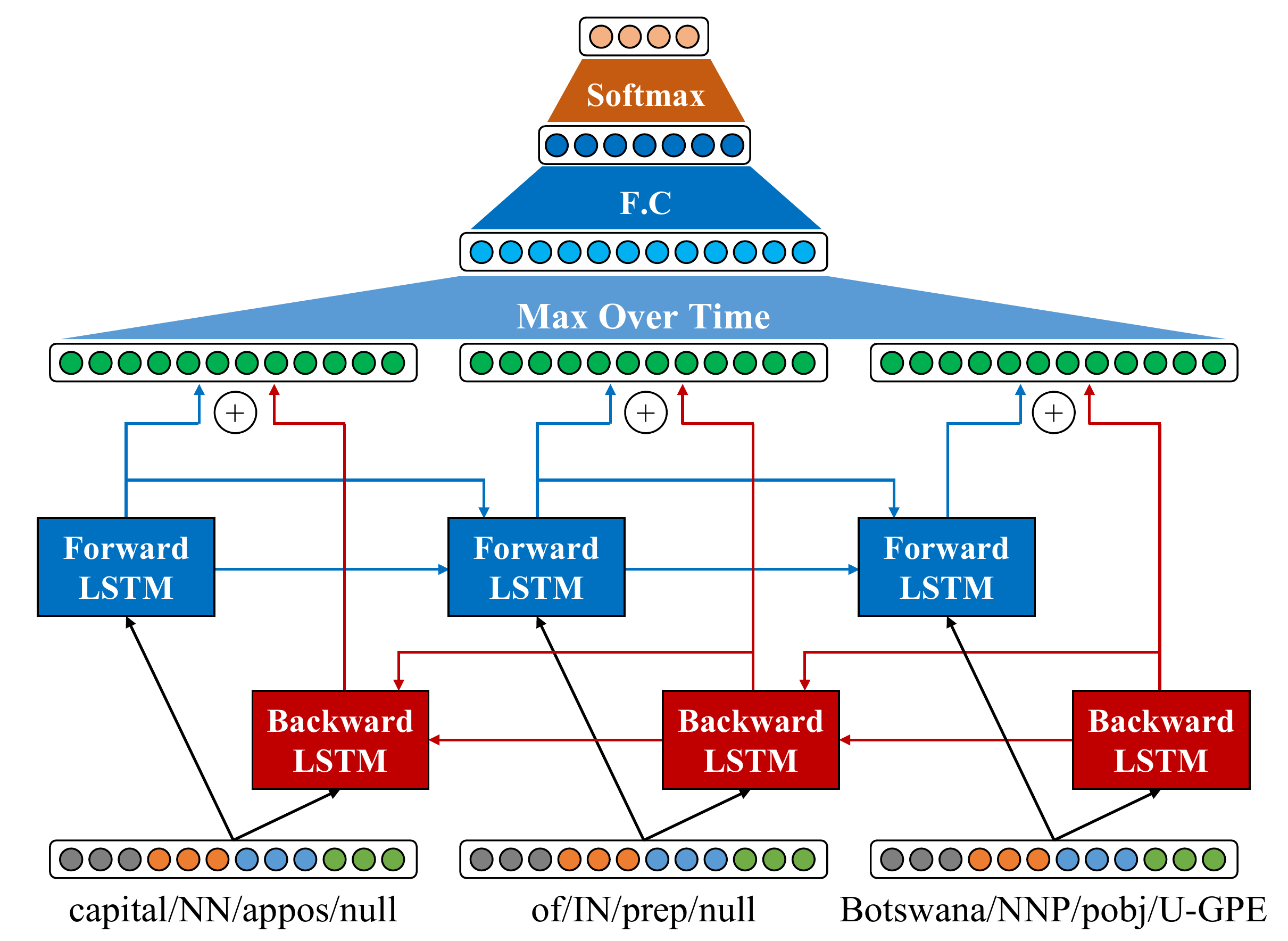}
    \caption{Neural network architecture for argument detection with input vectors from the \textit{path(capital, Botswana)} in the sentence, \textit{`The agency is located in Gaborone, capital of Botswana'}.}
    \label{fig:network}
\end{figure}
Figure \ref{fig:network} describes the architecture for the neural network used for argument detection. At each time-step, the network acquires an input vector from a node in \textit{path(rel, arg)}. The input vector is a concatenation of vectors from the following features: word, POS, dependency relation, and named entity.

Most of the deep learning applied NLP tasks leverage word embeddings trained with a large corpus in an unsupervised manner. The word embeddings capture the syntactic and semantic information of the words based on the context in the corpus. We pre-trained word embeddings from the English Wikipedia corpus with the skip-gram model in word2vec~\cite{Word2vec}. In doing so, we acquired the word embedding matrix, $M_{word} \in \mathbb{R}^{dim_{word}\times\mid W\mid}$, where $W$ is a set of words.

POS and dependency relations provide essential information regarding the syntactic structure of a sentence. However, there is no prevailing method for pre-training POS and dependency relation embeddings. In this work, we randomly initialized the POS and dependency relation embedding matrix, $M_{pos} \in \mathbb{R}^{dim_{pos}\times\mid P\mid}$ and $M_{dep} \in \mathbb{R}^{dim_{dep}\times\mid D\mid}$, where $P$ and $D$ are sets of POS tags and dependency labels, respectively. We then fine-tuned them in a supervised manner with back-propagation training.

Named entity recognition classifies each word into a pre-defined semantic category. We thus acquire the semantic types of words from the categories they belong to. Once again, the named entity embedding matrix, $M_{ne} \in \mathbb{R}^{dim_{ne}\times\mid N\mid}$ (where $N$ is a set of named entity tags), is randomly initialized and updated through back-propagation training.

The word, POS tag, dependency label, and named entity tag of the $t$-th node in \textit{path(rel, arg)} are associated with a vector, $word_t \in \mathbb{R}^{dim_{word}}$, $pos_t \in \mathbb{R}^{dim_{pos}}$, $dep_t \in \mathbb{R}^{dim_{dep}}$, and $ne_t \in \mathbb{R}^{dim_{ne}}$ in the embedding matrices, respectively. We concatenate these vectors to produce a single input vector of the $t$-th node, $x_t = [word_t, pos_t, dep_t, ne_t] \in \mathbb{R}^{dim_{word}+dim_{pos}+dim_{dep}+dim_{ne}}$.

A recurrent neural network (RNN) obtains the previous hidden state at each time-step, and creates and maintains the internal memory. By doing so, it can process arbitrary sequences of inputs. However, traditional RNNs have two well-known problems: vanishing and exploding gradients. If the input sequence is too long, the gradient can either decay or grow exponentially. An RNN with long short-term memory (LSTM) units was first introduced by~\newcite{LSTM} in order to tackle this problem with an adaptive gating mechanism. Among the many LSTM variants, we selected LSTM with peephole connections in the spirit of~\newcite{PeepHole}. Furthermore, we use both the forward and backward directional recurrent LSTM layer (see Appendix \ref{appendix:bi-directional}). This bi-directional architecture makes predictions based on information from both the past and the future. We obtain a bi-directional output vector $h_{t} \in \mathbb{R}^{dim_L}$ at each time-step from a vector sum of the forward ($h^{fw}_t$) and the backward ($h^{bw}_t$) LSTM layer output vectors.
{\begin{align}
    &h_t = h^{fw}_t+h^{bw}_t
\end{align}}%
We then convert an arbitrary number of bi-directional output vectors to a path-level feature vector $h_{path}$ through a max-over-time operation~\cite{SENNA}. This operation picks the salient features along the sequence of vectors to produce a single vector that is no longer related to the length of the sequence.
{\begin{align}
    &h_{path} = \max_{t}\{(h_t)_i\}\quad (0 \leq i \leq dim_L)
\end{align}}%
Subsequently, a fully connected layer non-linearly transforms the path-level feature vector to learn more complex features. We select the hyperbolic tangent activation function to obtain a higher-level feature vector $h_{higher} \in \mathbb{R}^{dim_H}$.
{\begin{align}
    &h_{higher} = \tanh(M_{higher}\cdot h_{path})
\end{align}}%
Finally, a softmax output layer projects $h_{higher}$ into a vector with dimensions equivalent to the number of classes. The softmax operation is then applied to obtain a vector $h_{out} \in \mathbb{R}^4$ with its elements representing the conditional probability for each class.
{\begin{align}
    &h_{out} = softmax(M_{out}\cdot h_{higher})
\end{align}}%

\section{Preposition Classification}
\label{sec:preposition_classification}
The neural network used for preposition classification is almost the same as the model used in the previous section. There are only two modifications to the preposition classification model. First, there is no penultimate fully connected layer in the model. We directly connect the max pooling layer to the softmax output layer. The second modification is to the number of output classes. The number of classes for preposition classification depends on the number of prepositions that appear in the positive samples. With 88 prepositions in the positive samples and one additional class for non-prepositions, the neural network model for preposition classification has $h_{out} \in \mathbb{R}^{89}$.

\section{Triple Extraction}
\label{sec:triple_extraction}
Triple extraction begins by aligning the prediction results as defined in the extraction template (see Table \ref{tbl:triple_extraction_template}). This alignment produces incomplete triples of arguments and relations that are incomplete phrases. We span the dependents of aligned words, \textit{arg1}, \textit{rel}, \textit{arg2}, and \textit{argN}, to ensure that the triples contain sufficient information from the sentences.

\begin{table}[ht]
    \centering
    \begin{adjustbox}{width=0.47\textwidth}
    \begin{tabular}{|c|l|}
        \hline
        \textbf{Relation type} & \textbf{Triple template} \\
        \hline
        Verb mediated & \textless [\textit{arg1}]; [\textit{rel}]; [\textit{arg2}]\textgreater \\
        \hline
        & \textless [\textit{arg2}]; be [\textit{rel}] [prep]; [\textit{argN}]\textgreater \\
        \hline
        & \textless [\textit{arg1}]; [\textit{rel}] [\textit{arg2}] [prep]; [\textit{argN}]\textgreater \\
        \hline
        Noun mediated & \textless [\textit{arg1}]; be [\textit{rel}] [prep]; [\textit{arg2}]\textgreater \\
        \hline
    \end{tabular}
    \end{adjustbox}
    \caption{Template for triple extraction.}
    \label{tbl:triple_extraction_template}
\end{table}

Previous Open IE systems assign a score for each extracted triple. The score is used to indicate the degree of correctness, since extracted triples are not always correct. We define a scoring function as below.
{\begin{align}
    &score(t) = dep(s)\times\frac{\sum\limits_{arg\in args} prob(arg)}{\mid args\mid}
\end{align}}%
where $t$ is an extracted triple from a sentence $s$, $dep(s)$ is the dependency parsing confidence score of $s$, $args$ is a set of arguments in $t$, and $prob(arg)$ is the conditional probability of $arg$ from the softmax output. Since errors in \textit{path(rel, arg)} are propagated to the final extraction, our scoring function is a mean of the conditional probabilities for arguments weighted by the dependency parsing confidence score of a sentence.

\section{Experiments}
\label{sec:experiments}
\subsection{Evaluation Settings}
We crawled news articles on the Web and randomly sampled 100 sentences for evaluation. Because Open IE extracts totally new relations from the sentences, there is no ground-truth set of extractions. For this reason, our natural choice for a performance metric was to calculate the precision over the number of extractions. This is a common metric in previous Open IE studies. The extractions were manually annotated for correctness and sorted according to their score, in descending order. We set our system to output extractions with scores over 0.75, in order to clarify our evaluation results.

\subsection{Comparison with State-of-the-Art Open IE Systems}
\begin{figure}[t]
    \centering
    \includegraphics[width=0.44\textwidth]{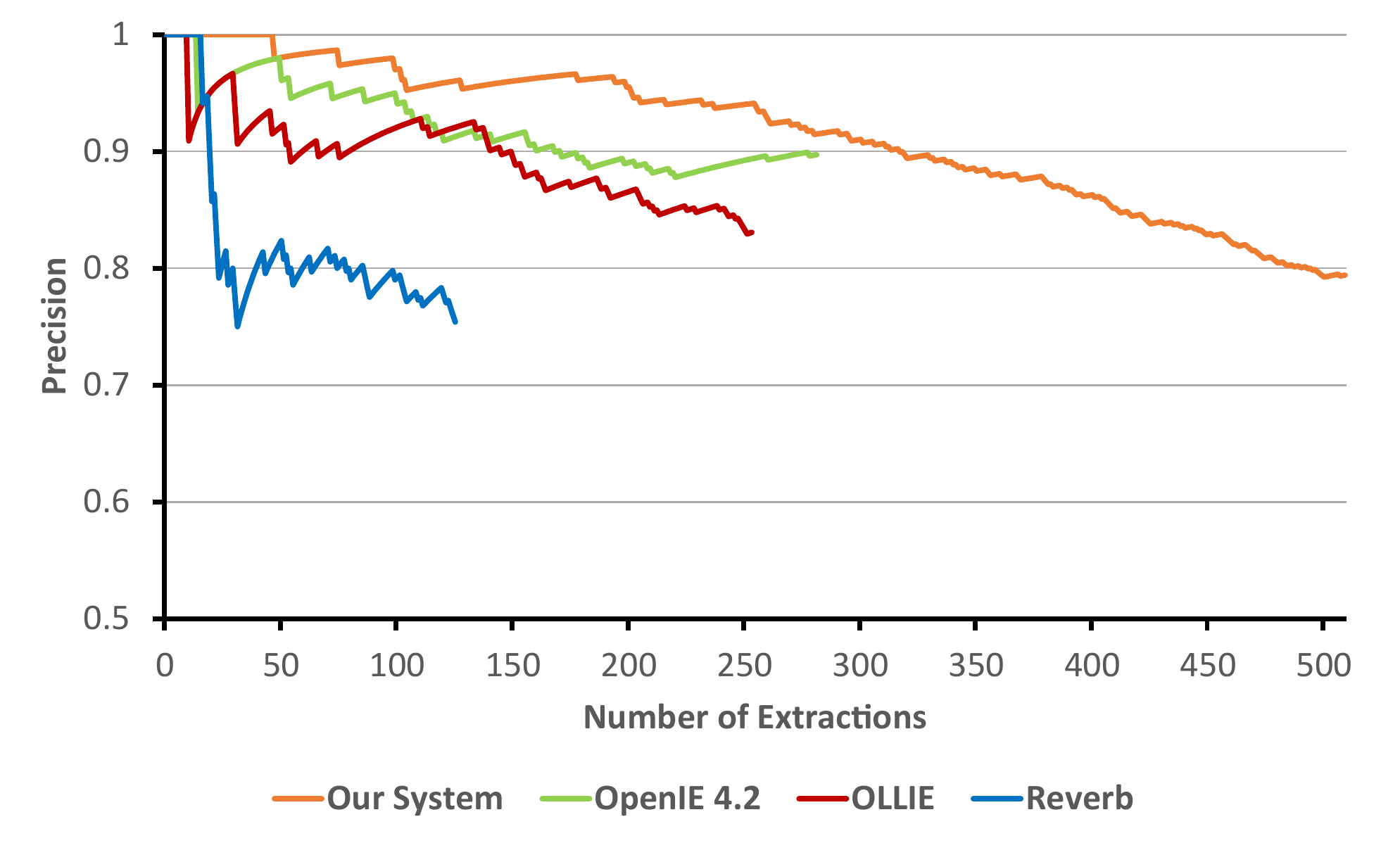}
    \caption{Our system produces more precise and abundant extractions than other state-of-the-art Open IE systems.}
    \label{fig:comparison_others}
\end{figure}
We compared our system with three widely used Open IE systems: Open IE 4.2\footnote{\url{https://github.com/knowitall/openie}}, OLLIE, and Reverb. Unlike Open IE 4.2 and OLLIE, our system does not determine whether the extractions are factual. Thus, we considered all extractions from Open IE 4.2 and OLLIE in the comparison without distinguishing the factuality of the extractions. Because there is no way to convert unary relations to binary relations, we discarded unary relations from Open IE 4.2. Our proposed system produced more extractions than the other Open IE systems, and it achieved the highest precision in all areas regarding the number of extractions (see Figure \ref{fig:comparison_others}). Specifically, the proposed system produced 1.62, 1.94, and 4.32 times more correct extractions than Open IE 4.2, OLLIE, and Reverb, respectively.

In addition to outperforming previous Open IE systems in terms of both precision and the total number of extractions, our system extracted implicit relations (see Appendix \ref{appendix:extraction_examples}). Extracting the implicit relations requires analyzing the context of the sentences, rather than merely setting boundaries to split the relations and arguments in sentences. Despite the relatively small proportion of implicit relations among correct extractions (3.8\%), they were indeed worth extracting, because they contributed to more abundant extractions. We compared our system to a model trained with samples without the augmented training set and found that these extractions were made from properly learning the relations from the augmented training set. All Open IE systems, apart from the proposed system, failed to extract implicit relations. Open IE 4.2 heuristically converted SRL outputs to produce most of its extractions. Because of its high reliance on SRL, it missed the implicit relations that SRL failed to capture. In a manner similar to the augmented training set, OLLIE automatically constructed training samples with seed-based distant supervision. However, OLLIE converted dependency paths connecting headwords of relations and arguments into pattern templates. Consequently, OLLIE failed to extract complex features from sentences. Reverb assumed that arguments and their relations appear consecutively in a sentence. Although this assumption is often correct, it is unsuitable when extracting implicit relations.

\subsection{Comparison with Different System Settings}
\begin{figure}[t]
    \centering
    \begin{subfigure}[b]{0.44\textwidth}
        \includegraphics[width=\textwidth]{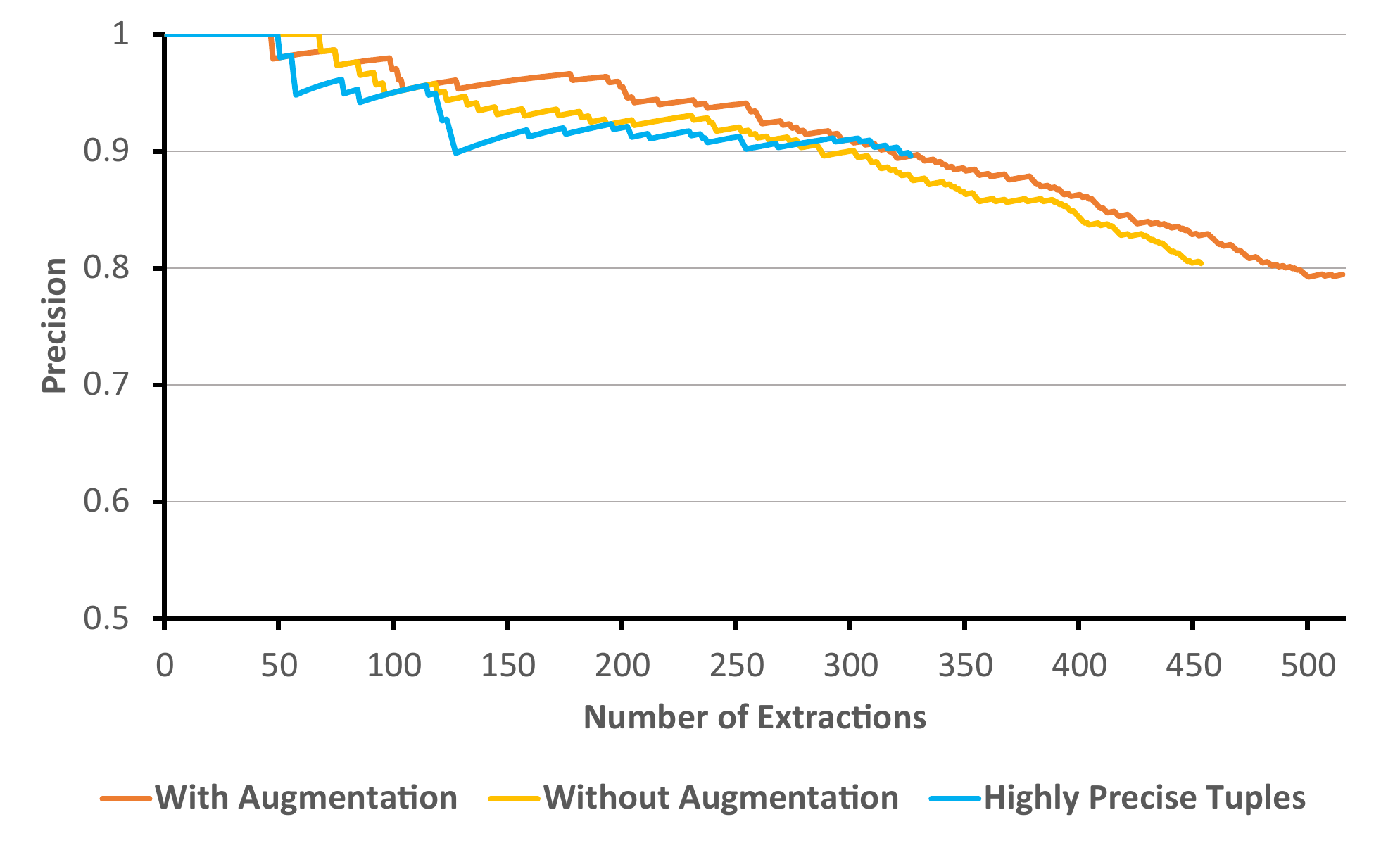}
        \caption{}
        \label{fig:comparison_settings_1}
    \end{subfigure}
    \begin{subfigure}[b]{0.44\textwidth}
        \includegraphics[width=\textwidth]{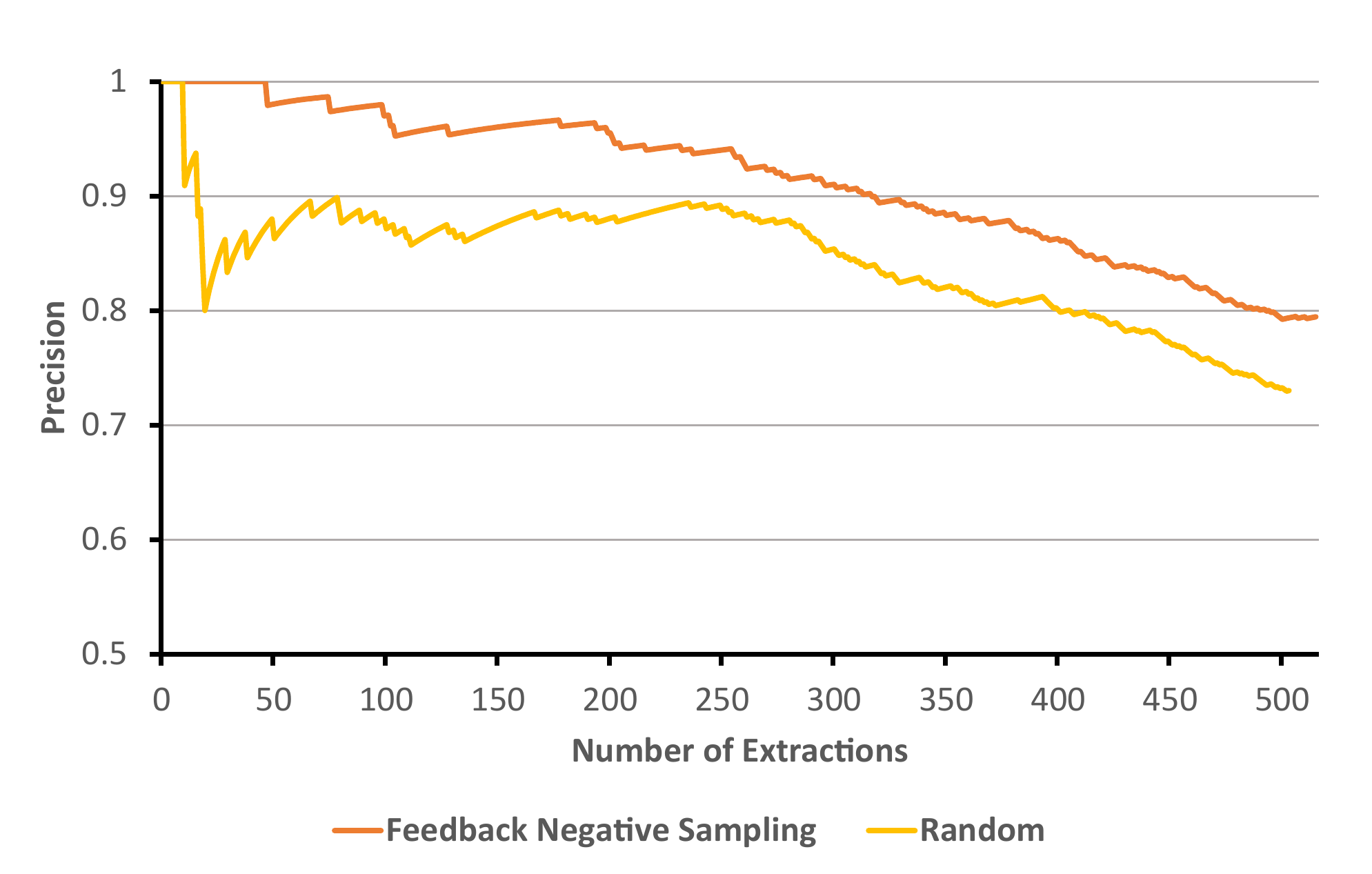}
        \caption{}
        \label{fig:comparison_settings_2}
    \end{subfigure}
    \begin{subfigure}[b]{0.44\textwidth}
        \includegraphics[width=\textwidth]{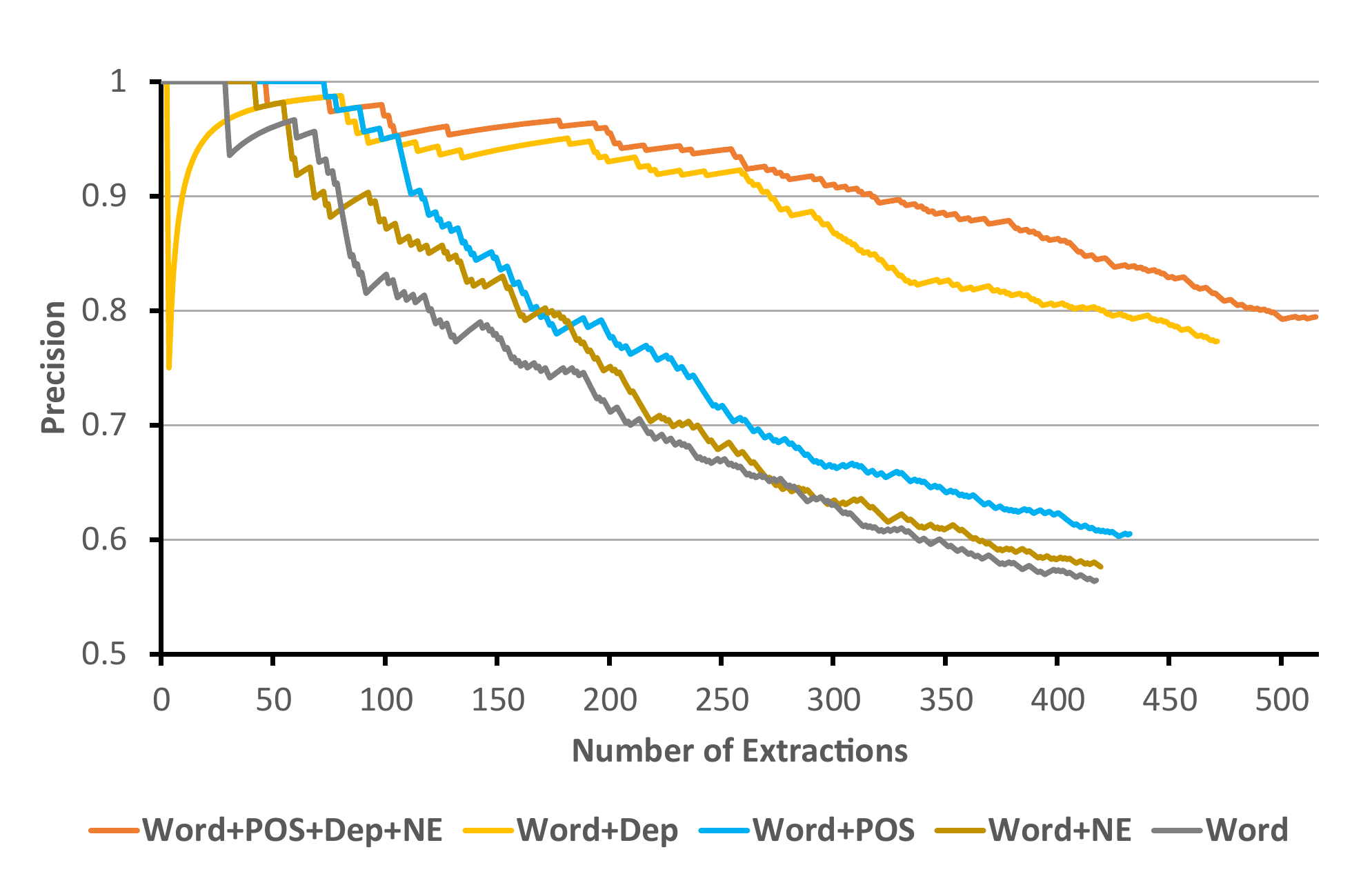}
        \caption{}
        \label{fig:comparison_settings_3}
    \end{subfigure}
    \caption{The best result is achieved with a model trained with positive samples from the augmented training set, negative samples from feedback negative sampling, and features from the word, POS, dependency relation, and named entity.}
    \label{fig:comparison_settings}
\end{figure}
Next, we analyzed how our system benefits from bi-directional LSTM networks (Figure \ref{fig:comparison_settings_1}). We compared two sets of extractions: extractions from a model trained with samples from highly precise tuples (\textit{Without Augmentation}), and extractions from a method using highly precise tuple extraction (\textit{Highly Precise Tuples}). The former set contained 1.25 times more correct extractions than the latter set. Moreover, when quality extractions were considered, the first set contained extractions that were more precise.

We analyzed the effect of augmenting the training set by comparing two models: a model trained with the augmented samples (\textit{With Augmentation}), and samples from the highly precise tuples (\textit{Without Augmentation}). Figure \ref{fig:comparison_settings_1} shows that augmented training set contributed to the production of 1.12 times more correct extractions with a slight boost in precision. Furthermore, the model without augmentation produced no extractions with implicit relations.

We analyzed the quality of samples from the augmented training set. Because the samples were from (sentence, seed triple) pairs, we manually checked whether in each pair the seed triple represented a valid relation in the sentence. Among the 200 randomly sampled pairs, 83.5\% were valid relations. Among pairs with invalid relations, 68\% were due to a failure in the distant supervision assumption, 29\% were due to errors in the seed triples, and 3\% were due to entity linking errors.

We compared two negative sampling strategies: feedback negative sampling (\textit{Feedback Negative Sampling}), and random sampling of non-positive paths (\textit{Random}) (see Figure \ref{fig:comparison_settings_2}). Feedback negative sampling achieved higher precision overall. The loss of precision from random sampling of non-positive paths was due to disagreements between the positive and negative samples.

We also analyzed how each input feature contributed to the extraction performance (see Figure \ref{fig:comparison_settings_3}). We set a baseline model with only the word feature (\textit{Word}) as the input. We then added the POS (\textit{Word+POS}), dependency relation (\textit{Word+Dep}), and named entity (\textit{Word+NE}) features one-by-one. Higher precision was achieved when the features were combined, compared to when only the word feature was used. Notably, the dependency relation feature boosted the precision considerably. By combining all four features (\textit{Word+POS+Dep+NE}), the precision further increased, with the added advantage of expanding the total number of correct extractions.

\subsection{Extraction Error Analysis}
We analyzed incorrect extractions and investigated the source of the errors. According to our analysis, 20\% of the errors were due to incorrect dependency parsing. Because the proposed system acquires a dependency path as an input, errors in dependency parsing were propagated throughout our system. Among the incorrect extractions without dependency parsing errors, 98\% of the errors were from argument detection, and 4\% were from preposition classification.

\section{Conclusion}
\label{sec:conclusion}
Our novel Open IE system with LSTM networks produced more precise and abundant extractions than state-of-the-art Open IE systems. In particular, the proposed system extracted implicit relations, unlike other Open IE systems. The advantages to the proposal stem from two contributions: a bi-directional recurrent architecture with LSTM units, enabling the extraction of higher-level features containing the contextual information in a sentence; and feedback negative sampling, which reduces the disagreements between positive and negative samples. To the best of our knowledge, this is the first work to apply deep learning to Open IE.

\bibliography{acl2016}
\bibliographystyle{acl2016}

\appendix

\section{Bi-Directional Recurrent Layer with LSTM Units}
\label{appendix:bi-directional}
We begin from the forward-directional recurrent layer with LSTM units that receive input sequences from beginning to end (Equations \ref{eq1}-\ref{eq6}).
\begin{align}
\begin{split}\label{eq1}
    f^{fw}_t = {}& \sigma(W^{fw}_f\cdot x_t+U^{fw}_f\cdot h_{t-1} \\
                 & +V^{fw}_f\cdot c^{fw}_{t-1}+b^{fw}_f)
\end{split}\\
\begin{split}\label{eq2}
    i^{fw}_t = {}& \sigma(W^{fw}_i\cdot x_t+U^{fw}_i\cdot h_{t-1} \\
                 & +V^{fw}_i\cdot c^{fw}_{t-1}+b^{fw}_i)
\end{split}\\
\begin{split}\label{eq3}
    g^{fw}_t = {}& \tanh(W^{fw}_g\cdot x_t+U^{fw}_g\cdot h_{t-1} \\
                 & +b^{fw}_g)
\end{split}\\
    c^{fw}_t = {}& i^{fw}_t\otimes g^{fw}_t+f^{fw}_t\otimes c^{fw}_{t-1}\label{eq4} \\
\begin{split}\label{eq5}
    o^{fw}_t = {}& \sigma(W^{fw}_o\cdot x_t+U^{fw}_o\cdot h_{t-1} \\
                 & +V^{fw}_o\cdot c^{fw}_t+b^{fw}_o)
\end{split}\\
    h^{fw}_t = {}& o^{fw}_t\otimes \tanh(c^{fw}_t)\label{eq6}
\end{align}
There are four components in the LSTM unit: a forget gate $f^{fw}_t$, an input gate $i^{fw}_t$, a candidate memory content $g^{fw}_t$, and an output gate $o^{fw}_t$. The forget and input gate receive the current input $x_t$, the previous output $h_{t-1}$, and the previous memory content $c^{fw}_{t-1}$. These are then multiplied with the matrices $W^{fw}$, $U^{fw}$, and $V^{fw}$, respectively. Then, the multiplied values are summed with a bias $b^{fw}$, and the result is non-linearly transformed through the sigmoid function $\sigma$ (Equations \ref{eq1}-\ref{eq2}). The candidate memory content receives the current input and the previous output, which are multiplied with the matrices $W^{fw}$ and $U^{fw}$, respectively. Then, the multiplied values are summed with a bias $b^{fw}$, and the result is non-linearly transformed through the hyperbolic tangent function $\tanh$ (Equation \ref{eq3}). The output gate also receives the current input and the previous output, but it considers the current memory content, rather than the previous memory content (Equation \ref{eq5}). The current memory content is a combination of candidate memory content and previous memory content, weighted by the values of the input gate and the forget gate, respectively (Equation \ref{eq4}). Finally, the current output is a normalized current memory content through the hyperbolic tangent function, weighted by the value of the output gate (Equation \ref{eq6}).

A potential problem with the forward LSTM layer is that it only considers information from the past. It thus fails to capture information from the future. We address this problem using an additional backward LSTM layer that receives input sequences from the end to the beginning (Equations \ref{eq7}-\ref{eq12}).
\begin{align}
\begin{split}\label{eq7}
    f^{bw}_t = {}& \sigma(W^{bw}_f\cdot x_t+U^{bw}_f\cdot h_{t+1} \\
                 & +V^{bw}_f\cdot c^{bw}_{t+1}+b^{bw}_f)
\end{split}\\
\begin{split}\label{eq8}
    i^{bw}_t = {}& \sigma(W^{bw}_i\cdot x_t+U^{bw}_i\cdot h_{t+1} \\
                 & +V^{bw}_i\cdot c^{bw}_{t+1}+b^{bw}_i)
\end{split}\\
\begin{split}\label{eq9}
    g^{bw}_t = {}& \tanh(W^{bw}_g\cdot x_t+U^{bw}_g\cdot h_{t+1} \\
                 & +b^{bw}_g)
\end{split}\\
    c^{bw}_t = {}& i^{bw}_t\otimes g^{bw}_t+f^{bw}_t\otimes c^{bw}_{t+1}\label{eq10} \\
\begin{split}\label{eq11}
    o^{bw}_t = {}& \sigma(W^{bw}_o\cdot x_t+U^{bw}_o\cdot h_{t+1} \\ 
                 & +V^{bw}_o\cdot c^{bw}_t+b^{bw}_o)
\end{split}\\
    h^{bw}_t = {}& o^{bw}_t\otimes \tanh(c^{bw}_t)\label{eq12}
\end{align}

\section{Training Details}
\label{appendix:training_details}
We set the prediction score threshold $p$ to 0.9 during feedback negative sampling. Furthermore, we set $dim_{word}$ to 300, and $dim_{pos}$, $dim_{dep}$, and $dim_{ne}$ to 50. Moreover, $dim_L$ was set to 450, which was equivalent to the dimensions of the input vector, and $dim_H$ was set to 50. Much like the regularization method used in~\newcite{RelationClassificationLSTM}, we assigned the input vector a dropout rate of 0.5. Because we apply a softmax operation for the final output, the natural choice for a training objective is cross-entropy.
{\begin{align}
    &J(\theta) = \sum\limits_{t\in T} \log p(y^{(t)}\mid x^{(t)}, \theta)
\end{align}}%
In the above equation, $T$ is a set of training samples, and $\theta$ = ($M_{pos}$, $M_{dep}$, $M_{ne}$, $M_{LSTM}$, $M_{higher}$, $M_{out}$) represents the network parameters, where $M_{LSTM}$ denotes the parameters in the LSTM units. We used the ADAM~\cite{ADAM} update rule to maximize the training objective through stochastic gradient descent over shuffled mini-batches. We set $\beta_1$ to 0.9, $\beta_2$ to 0.999, and $\epsilon$ to 1e-8 for the ADAM parameters.

\section{Extraction Examples}
\label{appendix:extraction_examples}
\begin{table*}[t]
    \centering
    \begin{adjustbox}{width=1\textwidth}
    \begin{tabular}{|c|l|c|c|}
        \hline
        \textbf{System} & \textbf{Extractions} & \multicolumn{2}{|c|}{\textbf{Annotation}}\\
        \hline
        
        \multicolumn{4}{|l|}{The UK Foreign Affairs Committee called upon Prime Minister David Cameron to boycott the event.} \\
        \hline
        our system & \makecell{\textit{\textless The UK Foreign Affairs Committee; called upon; Prime Minister David Cameron\textgreater}} & explicit & correct \\
        & \makecell{\textit{\textless The UK Foreign Affairs Committee; to boycott; the event\textgreater}} & explicit & correct \\
        & \makecell{\textit{\textless David Cameron; be Prime Minister of; UK\textgreater}} & implicit & correct \\
        & \makecell{\textit{\textless Prime Minister; called upon; David Cameron\textgreater}} & & incorrect \\
        \hline
        OpenIE 4.2 & \makecell{\textit{\textless The UK Foreign Affairs Committee; called; Prime Minister David Cameron\textgreater}} & explicit & correct \\
        & \makecell{\textit{\textless The UK Foreign Affairs Committee; called Prime Minister David Cameron to; boycott the event\textgreater}} & explicit & correct \\
        & \makecell{\textit{\textless The UK Foreign Affairs Committee; called to boycott; the event\textgreater}} & explicit & correct \\
        \hline
        OLLIE & \makecell{\textit{\textless The UK Foreign Affairs Committee; called upon; Prime Minister David Cameron\textgreater}} & explicit & correct \\
        & \makecell{\textit{\textless The UK Foreign Affairs Committee; to boycott; the event\textgreater}} & explicit & correct \\
        & \makecell{\textit{\textless The UK Foreign Affairs Committee; called to boycott; the event\textgreater}} & explicit & correct \\
        \hline
        Reverb & \makecell{\textit{\textless The UK Foreign Affairs Committee; called upon; Prime Minister David Cameron\textgreater}} & explicit & correct \\
        \hline
        
        \multicolumn{4}{|l|}{Article 7 of the UAE\textquotesingle s Provisional Constitution declares Islam the official state religion.} \\
        \hline
        our system & \makecell{\textit{\textless Article 7 of the UAE\textquotesingle s Provisional Constitution; declares; Islam the official state religion\textgreater}} & explicit & correct \\
        & \makecell{\textit{\textless Islam; be the official state religion of; the UAE\textgreater}} & implicit & correct \\
        & \makecell{\textit{\textless Islam; declares; the official state religion\textgreater}} & & incorrect \\
        \hline
        OpenIE 4.2 & \makecell{\textit{\textless Article 7 of the UAE\textquotesingle s Provisional Constitution; declares; Islam the official state religion\textgreater}} & explicit & correct \\
        \hline
        OLLIE & \makecell{No extractions found.} & & \\
        \hline
        Reverb & \makecell{\textit{\textless Article 7 of the UAE\textquotesingle s Provisional Constitution; declares; Islam\textgreater}} & explicit & correct \\
        \hline
        
        \multicolumn{4}{|l|}{It is 243 mi southeast of the capital Kiev on the Dnieper River, in the south-central part of Ukraine.} \\
        \hline
        our system & \makecell{\textit{\textless It; is in; the south central part of Ukraine\textgreater}} & explicit & correct \\
        & \makecell{\textit{\textless Kiev; be the capital of; Ukraine\textgreater}} & implicit & correct \\
        & \makecell{\textit{\textless It; be 243 mi southeast of the capital Kiev on; the Dnieper River\textgreater}} & explicit & correct \\
        & \makecell{\textit{\textless It; is; 243 mi southeast of the capital Kiev on the Dnieper River\textgreater}} & explicit & correct \\
        & \makecell{\textit{\textless It; be 243 mi southeast of on the Dnieper River of; the capital Kiev\textgreater}} & & incorrect \\
        \hline
        OpenIE 4.2 & \makecell{\textit{\textless It; is; 243 mi southeast of the capital Kiev on the Dnieper River, in the south-central part of Ukraine\textgreater}} & explicit & correct \\
        \hline
        OLLIE & \makecell{\textit{\textless It; is 243 mi southeast of the capital in; the south-central part of Ukraine\textgreater}} & explicit & correct \\
        & \makecell{\textit{\textless It; is 243 mi southeast of the capital on; the Dnieper River\textgreater}} & explicit & correct \\
        & \makecell{\textit{\textless It; is 243 mi of; the capital\textgreater}} & & incorrect \\
        & \makecell{\textit{\textless It; is; 243 mi southeast of the capital\textgreater}} & & incorrect \\
        & \makecell{\textit{\textless south-central; be part of; Ukraine\textgreater}} & & incorrect \\
        \hline
        Reverb & \makecell{\textit{\textless It; is; 243 mi\textgreater}} & & incorrect \\
        \hline
        
        \multicolumn{4}{|l|}{The gigantic 37m Merlion Statue, representing the mascot and national personification of Singapore, was prominently seen above the promenade.} \\
        \hline
        our system & \makecell{\textit{\textless The gigantic 37 m Merlion Statue; be seen; prominently\textgreater}} & explicit & correct \\
        & \makecell{\textit{\textless The gigantic 37 m Merlion Statue; be seen above; the promenade\textgreater}} & explicit & correct \\
        & \makecell{\textit{\textless The gigantic 37 m Merlion Statue; representing; the mascot and national personification of Singapore\textgreater}} & explicit & correct \\
        & \makecell{\textit{\textless The gigantic 37 m Merlion Statue; be seen; representing the mascot and national personification of} \\ \textit{Singapore\textgreater}} & explicit & correct \\
        & \makecell{\textit{\textless The gigantic 37 m Merlion Statue; be the mascot and national personification of; Singapore\textgreater}} & implicit & correct \\
        \hline
        OpenIE 4.2 & \makecell{\textit{\textless The gigantic 37 m Merlion Statue; representing; the mascot and national personification of Singapore\textgreater}} & explicit & correct \\
        & \makecell{\textit{\textless The gigantic 37 m Merlion Statue; was prominently seen above; the promenade\textgreater}} & explicit & correct \\
        \hline
        OLLIE & \makecell{\textit{\textless The gigantic 37 m Merlion Statue; was prominently seen above; the promenade\textgreater}} & explicit & correct \\
        \hline
        Reverb & \makecell{\textit{\textless the mascot; was prominently seen above; the promenade\textgreater}} & & incorrect \\
        \hline
        
        \multicolumn{4}{|l|}{The school was officially founded on August 22, 2014 when the NSHE Board of Regents approved a two year budget.} \\
        \hline
        our system & \makecell{\textit{\textless The school; be founded; officially\textgreater}} & explicit & correct \\
        & \makecell{\textit{\textless The school; be founded on; August 22 2014\textgreater}} & explicit & correct \\
        & \makecell{\textit{\textless the NSHE Board of Regents; approved; a two year budget\textgreater}} & explicit & correct \\
        & \makecell{\textit{\textless the NSHE Board of Regents; approved a two year budget on; August 22 2014\textgreater}} & implicit & correct \\
        & \makecell{\textit{\textless a two year budget; be approved on; August 22 2014\textgreater}} & implicit & correct \\
        & \makecell{\textit{\textless Regents; approved; a two year budget\textgreater}} & & incorrect \\
        & \makecell{\textit{\textless Regents; approved a two year budget on; August 22 2014\textgreater}} & & incorrect \\
        \hline
        OpenIE 4.2 & \makecell{\textit{\textless The school; was officially founded; when the NSHE Board of Regents approved a two year budget\textgreater}} & explicit & correct \\
        & \makecell{\textit{\textless The school; was officially founded on; August 22\textgreater}} & explicit & correct \\
        & \makecell{\textit{\textless the NSHE Board of Regents; approved; a two year budget\textgreater}} & explicit & correct \\
        \hline
        OLLIE & \makecell{\textit{\textless the NSHE Board of Regents; approved; a two year budget\textgreater}} & explicit & correct \\
        & \makecell{\textit{\textless The school; was officially founded; 2014\textgreater}} & & incorrect \\
        & \makecell{\textit{\textless The school; was officially founded 2014 when the NSHE Board of Regents approved a two year budget} \\ \textit{on; August 22\textgreater}} & & incorrect \\
        & \makecell{\textit{\textless The school; was officially founded 2014 when the NSHE Board of Regents approved a two year budget} \\ \textit{in; August 22\textgreater}} & & incorrect \\
        \hline
        Reverb & \makecell{\textit{\textless The school; was officially founded on; August 22 , 2014\textgreater}} & explicit & correct \\
        & \makecell{\textit{\textless the NSHE Board of Regents; approved; a two year budget\textgreater}} & explicit & correct \\
        \hline
    \end{tabular}
    \end{adjustbox}
    \caption{Our system extracts implicit relations missed by Open IE 4.2, OLLIE, and Reverb.}
\end{table*}
\end{document}